\documentclass[conference]{IEEEtran}
\IEEEoverridecommandlockouts

\usepackage{cite}
\usepackage{amsmath,amssymb,amsfonts}
\usepackage{algorithmic}
\usepackage{graphicx}
\usepackage{textcomp}
\usepackage{xcolor}
\usepackage{url}
\def\BibTeX{{\rm B\kern-.05em{\sc i\kern-.025em b}\kern-.08em
    T\kern-.1667em\lower.7ex\hbox{E}\kern-.125emX}}

\begin{document}

\title{A Conservative OCR-Enabled Workflow for R214 Sodium Screening of South African Packaged Foods%
\thanks{ISBN 979-8-3195-1782-1/26/\$31.00~\textcopyright{}2026 IEEE}
}

\author{
\IEEEauthorblockN{Mayimunah Nagayi}
\IEEEauthorblockA{
\textit{Department of Computer Science} \\
\textit{University of the Western Cape} \\
Cape Town, South Africa \\
4163113@myuwc.ac.za
}
\and
\IEEEauthorblockN{Alice Scaria Khan}
\IEEEauthorblockA{
\textit{School of Public Health} \\
\textit{University of the Western Cape} \\
Cape Town, South Africa \\
askhan@uwc.ac.za
}
\and
\IEEEauthorblockN{Tamryn Frank}
\IEEEauthorblockA{
\textit{School of Public Health} \\
\textit{University of the Western Cape} \\
Cape Town, South Africa \\
tfrank@uwc.ac.za
}
\and
\IEEEauthorblockN{Rina Swart}
\IEEEauthorblockA{
\textit{Department of Dietetics and Nutrition} \\
\textit{University of the Western Cape} \\
Cape Town, South Africa \\
rswart@uwc.ac.za
}
\and
\IEEEauthorblockN{Clement Nyirenda}
\IEEEauthorblockA{
\textit{Department of Computer Science and eResearch Office} \\
\textit{University of the Western Cape} \\
Cape Town, South Africa \\
cnyirenda@uwc.ac.za
}
}

\maketitle

\begin{abstract}
Using food package images to monitor sodium and salt content against South Africa's R214 sodium limits is challenging when screening decisions require product identity, nutrition facts panel evidence, reporting basis, and category-specific thresholds. This study presents a conservative image-based workflow that combines region detection, optical character recognition (OCR), product identity and sodium evidence extraction, R214 category assignment, deterministic threshold comparison, and independent vision language model comparison. The evaluation used 442 packaged food products and 3\,929 full package images from a real-world South African food packaging dataset. A YOLO26s small detector generated 4\,195 region crops, and strict post-processing produced one sodium evidence row per product. The integrated workflow produced 290 OUTSIDE R214 SCOPE, 139 REVIEW, seven SCREEN-PASS, and six SCREEN-FAIL outcomes. The independent Qwen2.5-VL 7B vision language model workflow produced 387 OUTSIDE R214 SCOPE, 31 REVIEW, twenty SCREEN-PASS, and four SCREEN-FAIL outcomes. The workflows agreed on exact R214 category assignment for 415 of 442 products (93.9\%) and on whether the assigned category was within R214 scope for 416 of 442 products (94.1\%). Final screening outcome agreement was 307 out of 442 products, or 69.5\%. Manual verification on 60 products showed lower strict outcome agreement than regulated status agreement, while all manual INSUFFICIENT DATA cases were kept out of SCREEN-PASS and SCREEN-FAIL by both automated workflows. The findings show that conservative image-based screening can organise package evidence, identify clear cases, and assign uncertain cases to REVIEW rather than forcing SCREEN-PASS or SCREEN-FAIL decisions.
\end{abstract}


\section{Introduction}

Food package labels are an important source of product-level nutrition, ingredient, and marketing information. Nutrition facts panels, ingredient lists, product names, quantity declarations, and other label elements are used in food composition databases, consumer tools, nutrition research, and food supply monitoring systems \cite{flip,foodswitch}. The Food Label Information Program (FLIP) shows how product-level label information can be structured for monitoring and research \cite{flip}, while FoodSwitch shows how branded food data and barcode scanning can support consumer nutrition information \cite{foodswitch}. In image-based work, this information must be recovered from package photographs, where text may be small, dense, multilingual, curved, or visually mixed with other package content \cite{nagayi_ocr,grocery_review}. South African studies based on package photographs also show that printed label evidence can support regulation-related assessment \cite{abdoolkarim_package}. For products that are not already available in a structured database, the physical package remains evidence that must be read and checked. This makes automated information extraction from food package images useful for regulation-related screening.

High sodium intake is associated with raised blood pressure and increased risk of cardiovascular disease \cite{who_sodium}. South Africa introduced mandatory sodium limits through Regulation R214, which sets maximum total sodium values per 100 g foodstuff for regulated processed food categories \cite{r214}. The limits are category specific. This means that sodium screening cannot be done by reading a sodium value alone. The product type must be identified, the correct regulatory category must be assigned, the sodium value and its basis must be checked, and the value must then be compared with the correct category threshold \cite{r214, charlton_compliance}. This makes R214 screening a structured evidence problem, not only a text extraction problem.

Prior South African sodium monitoring work has shown that R214 assessment depends on correct product categorisation, comparable sodium values, and expert checking \cite{r214, charlton_compliance, vdw_compliance}. Manual, laboratory-based, and database-based approaches remain important, but they are difficult to scale across large image collections and repeated market monitoring. Package images also create technical challenges, as useful label evidence may appear across several views and may be affected by small text, glare, curved surfaces, dense tables, multilingual text, and partial occlusion \cite{nagayi_ocr, grocery_review}. Optical character recognition (OCR) and extraction methods based on large language models can recover useful label text \cite{nagayi_ocr, assiri_llm}, but R214 screening still requires sodium or salt evidence, reporting basis checks, product type, category assignment, and threshold comparison \cite{r214, charlton_compliance}. In this study, conservative screening means that the workflow assigns REVIEW when product identity, category evidence, sodium or salt evidence, or reporting basis is unclear, rather than forcing a SCREEN-PASS or SCREEN-FAIL decision from incomplete evidence.

This paper proposes and evaluates a conservative image-based workflow for assessing packaged food products against official R214 sodium limits. The workflow links package evidence extraction, product identity evidence, sodium category classification, sodium and salt evidence checks, basis checking, and deterministic decision logic, with an independent Qwen2.5-VL 7B vision language model comparison used as a separate screening view.  Each product is assigned a conservative screening outcome, with the outcome definitions and decision rules described in Section~IV. The evaluation-set size was determined by the contents of three selected 2023 dataset folders, as described in Section~III-A.

\begin{enumerate}
    \item An image-based sodium screening workflow that combines package region detection, OCR, sodium and salt evidence selection, product identity extraction, R214 category classification, basis checking, and threshold comparison.

    \item A conservative decision framework that assigns products to SCREEN-PASS, SCREEN-FAIL, REVIEW, or OUTSIDE R214 SCOPE under the R214 screening logic without forcing unclear cases into SCREEN-PASS or SCREEN-FAIL decisions.

    \item An evaluation on 442 real-world South African packaged food products and 3\,929 package images, reporting evidence recovery, category assignment, and conservative screening outcomes.

    \item A comparison with an independent vision language model workflow and targeted manual checking to identify agreement patterns, disagreement types, and main failure sources.
\end{enumerate}

The rest of this paper is organised as follows. Section II reviews related work on sodium regulation monitoring, food package OCR, product recognition, food label databases, and vision language models. Section III describes the dataset and regulation context. Section IV presents the proposed methodology. Section V reports the results and evaluation. Section VI discusses the findings, limitations, and error sources. Section VII concludes the paper.

\section{Related Work}

\subsection{Sodium Regulation and Food Label Monitoring}

South African sodium regulation provides the policy context for this study. Regulation R214 defines maximum total sodium values per 100 g foodstuff for regulated processed food categories, with category-specific targets applied to product type rather than to sodium values alone \cite{r214}. Prior South African studies have assessed sodium content and industry compliance using processed food data, label-declared values, nutrition information panel data, laboratory analysis, and monitoring of regulated foodstuffs \cite{peters_sodium,swanepoel_sodium,vdw_compliance,charlton_compliance}. Together, these studies show that R214 screening requires correct product categorisation, comparable sodium values, and expert checking where category, panel values, and measurement basis must be interpreted together \cite{r214,charlton_compliance}. Food label databases, package-based systems, and South African package photograph studies also show that label evidence can support nutrition and regulation-related monitoring \cite{flip,foodswitch,abdoolkarim_package}. However, they rely mainly on structured, manual, or laboratory-supported data rather than automatically extracted package-image evidence.

\subsection{Food Package Text Extraction and Product Evidence}

Food package images are difficult to process automatically when labels appear on curved, reflective, multilingual, or visually crowded surfaces. Nagayi et al. \cite{nagayi_ocr} evaluated optical character recognition (OCR) performance on real-world South African food packaging labels and reported challenges linked to dense nutrition panels, ingredient text, multilingual content, glare, curved surfaces, and varied layouts. This work shows that OCR can recover useful label text, but text recovery alone does not create a regulation-ready screening decision. The recovered text still has to be linked to product identity, sodium or salt evidence, reporting basis, and the applicable R214 category. Guimarães et al. \cite{grocery_review} reviewed grocery label detection and recognition work and highlighted the difficulty of detecting and reading text from retail product packaging. Assiri et al. \cite{assiri_llm} studied extraction of nutritional elements and values from bilingual food labels using large language models, while Pettersson et al. \cite{pettersson_grocery} studied fine-grained grocery product recognition using product images and OCR text. These studies address important extraction and recognition tasks, but they do not combine food category assignment, per 100 g basis checking, sodium threshold comparison, and conservative review logic in one traceable screening workflow.

\subsection{Multimodal Models and Related Work Synthesis}

Vision language models (VLMs) and large language models (LLMs) are increasingly used for nutrition-related tasks. Ma et al. \cite{ma_vlm} integrated VLMs into a high-throughput nutrition screening workflow and showed that multimodal models can support faster processing of nutrition-related information. This work is relevant to package-based nutrition screening, but multimodal interpretation alone does not provide the controlled evidence checks and decision logic needed for regulation-specific screening. Across sodium monitoring, food label databases, package image text extraction, product recognition, and multimodal nutrition studies, existing work addresses important parts of the task, but the reviewed studies do not present a traceable image-based workflow that connects real-world package images to category assignment, reporting-basis checking, category-specific threshold comparison, and conservative screening outcomes. This gap motivates the workflow proposed in this paper.

\section{Dataset and Regulation Context}

\subsection{Dataset Source and Package Views}

This study uses a real-world image dataset from the Department of Dietetics and Nutrition at the University of the Western Cape. The dataset forms part of the South African Nutrition Facts Panel project. The images were collected in 2023 by trained personnel and accessed through the Department of Dietetics and Nutrition Research project site via SharePoint. The subset used in this study was drawn from three image folders: 20230608\_04\_03, 20230608\_04\_18, and 20230612\_04\_18. The folders were selected after initial inspection of 2023 folders for use as the study subset, without a formal random or category-based sampling procedure. The selected folders contained a range of packaged food products and were not filtered in advance to include only products from R214 categories. There were no duplicate products in the selected subset. Each product had a unique product identifier, and all products and their associated package images in the selected folders were included, giving 442 packaged food products and 3\,929 full package images. The study size was therefore determined by the contents of the selected folders rather than by a predefined sample-size target. Each product was represented by multiple package views rather than a single image. The number of images per product ranged from three to twenty-seven, with an average of 8.9 images per product. This multi-view structure was important for R214 screening, as product identity, sodium values, reporting basis information, ingredients, and quantity declarations may appear on different parts of the package. These package views therefore provided the visual evidence used for product identity extraction, sodium and salt evidence extraction, category assignment, and screening.

\subsection{R214 Sodium Screening Context}

Regulation R214 defines maximum total sodium values per 100~g foodstuff for thirteen regulated processed food categories \cite{r214}. Since the package images used in this study were collected in 2023, the workflow used the 2019 R214 limits for screening. The official R214 categories were used as the regulatory category space. The working label space used codes 1 to 13 for the official R214 sodium categories and an operational code 14 for products outside the R214 sodium category scope based on the available evidence. Code 14 is not an official R214 foodstuff category. An OUTSIDE R214 SCOPE outcome therefore indicates that the available evidence did not support assignment to one of the thirteen R214 categories, not that sodium was absent from the product. Products with unclear identity or unclear category evidence were assigned to REVIEW during the decision stage rather than being forced into code 14. R214 screening therefore requires both category evidence and comparable sodium evidence, as described in the methodology section.

\begin{figure*}[!t]
    \centering
    \includegraphics[
        width=0.75\textwidth,
        height=0.70\textheight,
        keepaspectratio
    ]{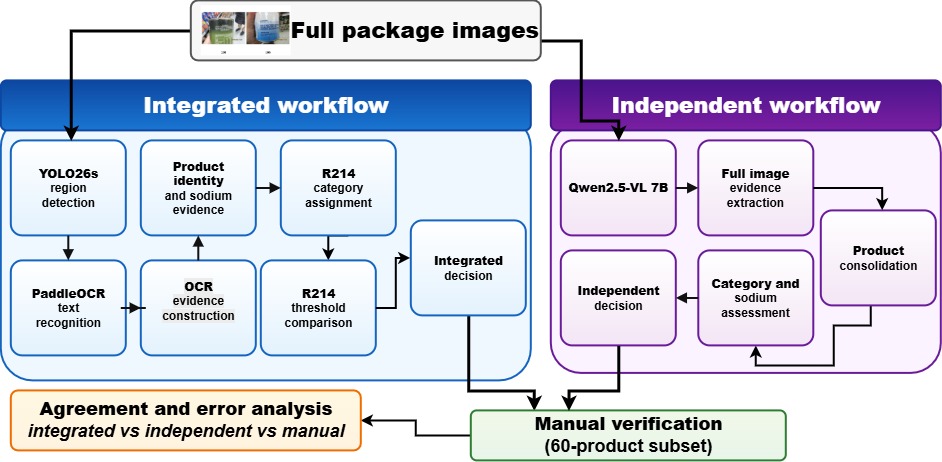}
    \caption{Overview of the image-based R214 sodium screening and validation workflow. The integrated workflow produces product-level screening outcomes, the independent workflow provides a separate comparison, and manual verification checks a 60 product subset for agreement and error analysis.}
    \label{fig:proposed_workflow}
\end{figure*}

\section{Methodology}

This section describes the staged image-based workflow used to convert full package images into product-level R214 sodium screening outcomes. As shown in Fig.~\ref{fig:proposed_workflow}, the study used an integrated workflow, an independent workflow, and manual verification. The integrated workflow combined region detection, crop generation, OCR, evidence construction, product identity extraction, sodium and salt evidence extraction, R214 category assignment, and deterministic threshold comparison. The independent workflow used full package images with Qwen2.5-VL 7B, a vision-language model from the Qwen2.5-VL model family, to provide a separate screening view \cite{qwen25vl}. Manual verification was used on a fixed 60 product subset to check agreement patterns and error cases.

\subsection{Region Detection and Crop Generation}

Detector training used an NVIDIA GeForce RTX 3070 graphics processing unit (GPU) with Ultralytics 8.4.47 while the full study workflow ran on the University of the Western Cape high performance computing (HPC) cluster using Slurm GPU jobs on an NVIDIA L4 GPU node with four CPU cores and 48 GB memory. The computer vision stage used a YOLO26s small detector from the You Only Look Once (YOLO) model family, implemented with the Ultralytics YOLO software \cite{ultralytics_yolo}, to locate four evidence regions: nutrition facts panel, ingredients, product name, and quantity. The detector was trained on a separate labelled image set from folder \texttt{20230614\_04\_18}, so the detector training images did not overlap with the 442-product study set. Annotations were prepared in Microsoft Visual Object Tagging Tool using four tags: \texttt{nfp}, \texttt{ingredients}, \texttt{Product\_name}, and \texttt{Quantity} \cite{microsoft_vott}. After a product-based split, the labelled dataset contained 176 products and 718 images: 141 products (80.1\%) for training, eighteen products (10.2\%) for validation, and seventeen products (9.7\%) for testing. Training used YOLO26s pretrained weights, image size 896, batch size 4, 150 maximum epochs, patience 30, and seed 42. Inference on the 3\,929 study images used image size 896, confidence threshold 0.35, and intersection over union threshold 0.60, with class-specific padding before crop saving.

\subsection{Text Recognition and Evidence Construction}

Text recognition was applied to the detected region crops using PaddleOCR 3.5.0, \texttt{paddlepaddle-gpu} 3.3.0, and the \texttt{latin\_PP-OCRv5\_mobile\_rec} recognition model. Previous OCR accuracy evaluation against manually transcribed package text from the same broader image collection is reported by Nagayi et al.~\cite{nagayi_ocr}. Text recognition was run on the HPC cluster using GPU resources. Recognised text was cleaned by normalising Unicode, removing hidden characters, replacing newlines with spaces, and collapsing repeated spaces. Weak crop-level outputs were retried with an enhanced crop variant, and the retry output was retained only when it improved the candidate score. Crop-level outputs were grouped by product and region to build product-level evidence. Region candidates were scored using task-specific cues. Nutrition facts panel candidates were scored using nutrition terms, sodium or salt terms, numeric content, per 100 g or per 100 ml cues, and table-like content. Ingredients, product name, and quantity candidates were scored using region-specific text patterns. Nutrition facts panel candidates were retained for sodium and salt parsing, and the strict evidence step selected one sodium evidence row per product.

\subsection{Product Identity, Sodium Evidence, and Category Assignment}

Product identity was extracted from full package images using Qwen2.5-VL 7B through Ollama, with the model identifier \texttt{qwen2.5vl:7b}. Full images were used for this step to capture stylised and brand-led product names, front-of-pack text, visual context, and multiple package views that OCR crops alone may miss. The model returned product name, product type, quantity, visible brand, visible text, and a short reason. Image-level outputs were scored, and the highest-scoring candidate was selected as product-level identity evidence; low-scoring or incomplete outputs were flagged for review.

Sodium and salt evidence was extracted from nutrition facts panel candidate text. The parser searched for sodium and salt labels, nearby numeric values, mg or g units, and the order of per 100~g and serving information. Values were converted to sodium in mg per 100~g where supported. Strict automatic evidence was limited to direct sodium in mg per 100~g or salt in g per 100~g with clear column order and no quality-assurance flag. Other conversions or unclear cases were assigned to REVIEW. Salt conversion used standard sodium--salt conversion factors \cite{who_sodium_reduction}. Products without usable values, comparable units, or a clear per 100 g basis were assigned to REVIEW.

The R214 category assignment used product-level identity evidence. The classifier used strict rules first, followed by a Groq API assisted fallback with \texttt{llama-3.3-70b-versatile} at temperature 0 for uncertain cases involving brand-specific or non-standard product wording \cite{groq_api}. The category space used official R214 categories 1 to 13 and operational code 14 for products not applicable to R214 based on available evidence. Products with unclear category evidence were assigned to category review rather than being forced into code 14.

\subsection{Screening Outcomes, Decision Rules, and Evaluation Measures}

The study used four conservative screening outcomes. These outcomes were used for screening and prioritisation, not for final legal determination.

\begin{itemize}
    \item \textbf{SCREEN-PASS.} The available evidence supported an applicable R214 category, comparable sodium evidence, and a value at or below the relevant R214 limit.

    \item \textbf{SCREEN-FAIL.} The available evidence supported an applicable R214 category, comparable sodium evidence, and a value above the relevant R214 limit.

    \item \textbf{REVIEW.} The available evidence was incomplete, unclear, conflicting, or not reliable enough for a SCREEN-PASS or SCREEN-FAIL decision. This included unclear product identity, uncertain category assignment, weak sodium or salt extraction, unclear reporting basis, or non-comparable units.

    \item \textbf{OUTSIDE R214 SCOPE.} The available product evidence did not support assignment to an applicable R214 sodium category. In this study, OUTSIDE R214 SCOPE is only an R214 screening outcome, not a statement about all possible food labelling or food composition regulations.
\end{itemize}

The final decision step combined category evidence, sodium evidence, and the 2019 R214 limits. If category evidence required review, the final decision was REVIEW. If a product was assigned operational code 14 without category review, the final decision was OUTSIDE R214 SCOPE. If a product was assigned an R214 category but sodium evidence was not READY, the final decision was REVIEW. If a product had both an applicable R214 category and READY sodium evidence, the sodium value was compared with the relevant 2019 R214 limit. Values at or below the limit were assigned SCREEN-PASS, and values above the limit were assigned SCREEN-FAIL. Additional safety checks flagged laboratory-method text, extreme or implausibly low sodium values, unclear column order, unusual salt conversions, and uncertain category assignments. Where comparison was possible but further checking was still needed, provisional SCREEN-PASS or SCREEN-FAIL labels were retained for audit, while these products were reported as REVIEW.

Detector performance was reported using precision, recall, and mean average precision (mAP). Agreement between automated workflows was measured using category agreement, regulated status agreement, final decision group agreement, and direct SCREEN-PASS versus SCREEN-FAIL conflict. Category agreement counted products where both workflows assigned the same R214 category or the same outside-scope grouping. Regulated status agreement counted products where both workflows agreed on whether the product was assigned to an applicable R214 sodium category. Final decision group agreement compared the four reported screening outcomes directly. Direct SCREEN-PASS versus SCREEN-FAIL conflict counted cases where one workflow assigned SCREEN-PASS and the other assigned SCREEN-FAIL.

\subsection{Independent Workflow and Manual Verification}

The independent workflow used Qwen2.5-VL 7B, a vision language model run through Ollama at temperature 0, on full package images only, and did not use integrated workflow outputs such as YOLO crops, PaddleOCR text, selected sodium evidence, category assignments, or final decisions. Qwen performed image evidence extraction, product consolidation, R214 category assessment, sodium assessment, and final screening decision, while Python handled execution, structured-output parsing, validation checks, retries, and result logging. Generic category rules were used to reduce forced assignment of out-of-scope products into official R214 categories. SCREEN-PASS and SCREEN-FAIL decisions were kept only when the sodium or salt evidence was internally valid and clearly comparable per 100 g; otherwise, the product was assigned to REVIEW.

Manual verification was conducted on a fixed subset of 60 products selected at product level. Products were matched across the manual file, integrated workflow output, and independent workflow output using the product identifier. Manual checking used the original package images, including evidence across package views where needed, and extracted evidence files to confirm product identity, R214 category, sodium or salt evidence, reporting basis, and screening outcome. The subset focused on important screening cases, including regulated or priority products, uncertain cases, and disagreement cases between the two automated workflows.

\section{Results and Evaluation}

This section reports the processing outputs, workflow screening outcomes, automated agreement, and manual verification results. Detector performance was evaluated using precision, recall, and mean average precision (mAP), while workflow agreement was evaluated using the product-level agreement measures defined in the methodology section.

\subsection{Processing and Evidence Recovery}

The detector provided usable localisation for the downstream R214 screening workflow. On the validation split, it achieved 0.744 precision, 0.693 recall, 0.718 mAP at 0.5, and 0.480 mAP at 0.5:0.95. On the held-out test split, it achieved 0.897 precision, 0.877 recall, 0.887 mAP at 0.5, and 0.638 mAP at 0.5:0.95. The nutrition facts panel class had the strongest test result, with 0.995 mAP at 0.5, while quantity was the weakest class, with 0.778 mAP at 0.5. The trained detector was applied to all 3\,929 full package images and generated 4\,195 region crops from 2\,901 images. PaddleOCR processed all crops with no processing errors. Product-level text evidence was constructed for all 442 products, with 952 nutrition facts panel candidates retained for sodium and salt parsing. The sodium evidence step produced one strict sodium evidence row per product. Of the 442 products, 207 were marked READY for automatic sodium comparison and 235 were assigned to REVIEW. Category assignment was rule-based for 302 products, used the Groq fallback for 49 products, and required category review for 91 products. Table~\ref{tab:processing_evidence_summary} summarises detector performance, crop recovery, and product-level evidence preparation.

\begin{table}[!htbp]
\caption{Processing and Evidence Recovery Summary}
\label{tab:processing_evidence_summary}
\centering
\scriptsize
\setlength{\tabcolsep}{3pt}
\renewcommand{\arraystretch}{1.08}
\begin{tabular}{p{0.68\columnwidth}r}
\hline
Measure & Result \\
\hline
\multicolumn{2}{l}{\textit{Detector performance}} \\
Validation precision / recall & 0.744 / 0.693 \\
Validation mAP@0.5 / mAP@0.5:0.95 & 0.718 / 0.480 \\
Test precision / recall & 0.897 / 0.877 \\
Test mAP@0.5 / mAP@0.5:0.95 & 0.887 / 0.638 \\
Nutrition facts panel test mAP@0.5 & 0.995 \\
Quantity test mAP@0.5 & 0.778 \\
\hline
\multicolumn{2}{l}{\textit{Image processing and crop recovery}} \\
Images processed / with crops & 3\,929 / 2\,901 \\
Total crops generated & 4\,195 \\
Product name / ingredients crops & 1\,461 / 1\,093 \\
Nutrition facts panel / quantity crops & 952 / 689 \\
\hline
\multicolumn{2}{l}{\textit{Evidence and category preparation}} \\
Strict sodium evidence rows & 442 \\
Sodium READY / REVIEW products & 207 / 235 \\
Salt converted products & 15 \\
Rejected sodium candidates & 150 \\
Category assignment: rule / Groq / review & 302 / 49 / 91 \\
\hline
\end{tabular}
\end{table}

\subsection{Workflow Outcomes and Automated Agreement}

The integrated workflow produced six detailed decision labels. For reporting and comparison with the independent workflow, provisional SCREEN-PASS and SCREEN-FAIL labels were retained for audit but reported as REVIEW when further checking was required. The integrated workflow therefore assigned 290 products to OUTSIDE R214 SCOPE, 139 to REVIEW, seven to SCREEN-PASS, and six to SCREEN-FAIL. The independent Qwen2.5-VL 7B workflow assigned 387 products to OUTSIDE R214 SCOPE, 31 to REVIEW, twenty to SCREEN-PASS, and four to SCREEN-FAIL. Agreement values were calculated at product level across all 442 products. Category agreement between the two automated workflows was 415 out of 442 products, or 93.9\%. Regulated status agreement was 416 out of 442 products, or 94.1\%. Final decision group agreement was 307 out of 442 products, or 69.5\%. One direct SCREEN-PASS versus SCREEN-FAIL conflict was found between the integrated and independent workflows. Table~\ref{tab:workflow_outcomes_agreement} summarises the reported outcomes and automated agreement.

\begin{table}[!htbp]
\caption{Workflow Outcomes and Automated Agreement}
\label{tab:workflow_outcomes_agreement}
\centering
\scriptsize
\setlength{\tabcolsep}{4pt}
\renewcommand{\arraystretch}{1.08}
\begin{tabular}{lcc}
\hline
Measure & Integrated & Independent \\
\hline
\multicolumn{3}{l}{\textit{Reported screening outcomes}} \\
OUTSIDE R214 SCOPE & 290 & 387 \\
REVIEW & 139 & 31 \\
SCREEN-PASS & 7 & 20 \\
SCREEN-FAIL & 6 & 4 \\
Total & 442 & 442 \\
\hline
\multicolumn{3}{l}{\textit{Agreement between automated workflows}} \\
Category agreement & \multicolumn{2}{c}{415/442 (93.9\%)} \\
Regulated status agreement & \multicolumn{2}{c}{416/442 (94.1\%)} \\
Final decision group agreement & \multicolumn{2}{c}{307/442 (69.5\%)} \\
Direct SCREEN-PASS versus SCREEN-FAIL conflict & \multicolumn{2}{c}{1} \\
\hline
\end{tabular}
\end{table}

\subsection{Manual Verification}

Manual verification was conducted on a fixed, targeted subset of 60 products selected at product level. Manual review assigned twenty-six products to SCREEN-PASS, twenty-one to SCREEN-FAIL, eight to INSUFFICIENT DATA, and five to OUTSIDE R214 SCOPE. For the same products, the integrated workflow assigned seven to SCREEN-PASS, six to SCREEN-FAIL, and 47 to REVIEW. The independent workflow assigned sixteen to SCREEN-PASS, four to SCREEN-FAIL, 24 to REVIEW, and sixteen to OUTSIDE R214 SCOPE. The 60-product subset covered 20 of the 24 SCREEN-PASS/SCREEN-FAIL products in the independent workflow, including all four independent SCREEN-FAIL products. Manual verification agreement was calculated at product level across the subset. Strict outcome agreement with manual review was 12 out of 60 products, or 20.0\%, for the integrated workflow, and 20 out of 60 products, or 33.3\%, for the independent workflow. Of the thirteen SCREEN-PASS or SCREEN-FAIL decisions retained by the integrated workflow after conservative review, twelve agreed with the manual screening direction. Regulated status agreement was 55 out of 60 products, or 91.7\%, for manual versus the integrated workflow, and 47 out of 60 products, or 78.3\%, for manual versus the independent workflow. All eight manual INSUFFICIENT DATA cases were kept out of SCREEN-PASS and SCREEN-FAIL by both automated workflows. Direct SCREEN-PASS versus SCREEN-FAIL conflicts were limited to one case for manual versus the integrated workflow and three cases for manual versus the independent workflow. Table~\ref{tab:manual_verification_checks} summarises the manual verification checks.

\begin{table}[!htbp]
\caption{Manual Verification Checks on the 60 Product Subset}
\label{tab:manual_verification_checks}
\centering
\scriptsize
\setlength{\tabcolsep}{4pt}
\renewcommand{\arraystretch}{1.08}
\begin{tabular}{@{}p{0.53\columnwidth}cc@{}}
\hline
Check & Integrated & Independent \\
\hline
Strict agreement with manual & 12/60 (20.0\%) & 20/60 (33.3\%) \\
Regulated status agreement & 55/60 (91.7\%) & 47/60 (78.3\%) \\
INSUFFICIENT DATA not SCREEN-PASS/SCREEN-FAIL & 8/8 & 8/8 \\
Direct SCREEN-PASS versus SCREEN-FAIL conflict & 1 & 3 \\
\hline
\end{tabular}
\end{table}

\section{Discussion}

The results support the conservative decision design described in Section~IV. The detector recovered useful package regions, with stronger nutrition facts panel localisation than quantity localisation, but downstream uncertainty remained after text recognition, sodium parsing, basis checking, and category assignment. This shows that the main difficulty was not only reading label text, but converting package evidence into reliable product-level evidence for R214 screening. Relevant evidence could also occur across different package views and had to be linked to the correct reporting basis and regulatory category.

The comparison between workflows shows that broad R214 relevance was more stable than exact screening outcome. Category agreement and regulated status agreement were higher than final decision group agreement, while the integrated workflow assigned more products to REVIEW than the independent workflow. Products requiring further checking remained under REVIEW even when a threshold direction could be calculated. Manual verification showed the same conservative pattern. Of the thirteen SCREEN-PASS or SCREEN-FAIL decisions retained by the integrated workflow, twelve agreed with the manual screening direction, while manual checking could inspect evidence across package views where needed. All eight manual INSUFFICIENT DATA cases were kept out of SCREEN-PASS and SCREEN-FAIL by both automated workflows. The workflow is therefore best understood as a screening and prioritisation approach that retains clearer decisions and directs uncertain products to review.

\section{Conclusion}

This paper presented a conservative image-based workflow for screening packaged food products against South Africa's R214 sodium limits. On 442 products and 3\,929 full package images, the integrated workflow showed a more conservative decision pattern than the independent Qwen2.5-VL 7B workflow, with more products assigned to REVIEW and only one direct SCREEN-PASS versus SCREEN-FAIL conflict between the two automated workflows. Manual verification showed stronger alignment on regulated status than on exact screening outcome, and all manual INSUFFICIENT DATA cases were kept out of SCREEN-PASS and SCREEN-FAIL. The workflow is not intended to replace human judgement or official compliance assessment; its value is in organising package image evidence, identifying clear screening cases, and prioritising uncertain products for manual review. Future work should expand manual verification and extend the workflow to additional labelling regulations.

\section*{Acknowledgment}
Most computational processing in this study was performed using the University of the Western Cape's High Performance Computing and Research Cloud facilities (\url{https://eresearch.uwc.ac.za}).


\begin{thebibliography}{00}

\bibitem{flip}
M. Ahmed, A. Schermel, J. J. Lee, M. Weippert, B. Franco-Arellano, and M. R. L'Abb\'{e}, ``Development of the Food Label Information Program: a comprehensive Canadian branded food composition database,'' \emph{Frontiers in Nutrition}, vol. 8, Art. no. 825050, 2022, doi: \url{https://doi.org/10.3389/fnut.2021.825050}.

\bibitem{foodswitch}
E. Dunford, H. Trevena, C. Goodsell, K. H. Ng, J. Webster, A. Millis, S. Goldstein, O. Hugueniot, and B. Neal, ``FoodSwitch: a mobile phone app to enable consumers to make healthier food choices and crowdsourcing of national food composition data,'' \emph{JMIR mHealth and uHealth}, vol. 2, no. 3, Art. no. e37, 2014, doi: \url{https://doi.org/10.2196/mhealth.3230}.

\bibitem{nagayi_ocr}
M. Nagayi, A. S. Khan, T. Frank, R. Swart, and C. Nyirenda, ``Evaluating OCR performance on food packaging labels in South Africa,'' in \emph{Artificial Intelligence Research: 6th Southern African Conference, SACAIR 2025, Cape Town, South Africa, December 1--5, 2025, Proceedings}, A. Gerber and A. W. Pillay, Eds. Cham, Switzerland: Springer, 2026, pp. 127--143, doi: \url{https://doi.org/10.1007/978-3-032-11733-5_8}.

\bibitem{grocery_review}
V. Guimarães, J. Nascimento, P. Viana, and P. Carvalho, ``A review of recent advances and challenges in grocery label detection and recognition,'' \emph{Applied Sciences}, vol. 13, no. 5, Art. no. 2871, 2023, doi: \url{https://doi.org/10.3390/app13052871}.

\bibitem{abdoolkarim_package}
S. Abdool Karim, T. Frank, A. S. Khan, M. G. Tlhako, S. K. Joni, E. C. Swart, \emph{et al.}, ``An assessment of compliance with proposed regulations to restrict on package marketing of packaged foods to improve nutrition in South Africa,'' \emph{BMC Nutrition}, vol. 11, Art. no. 17, 2025, doi: \url{https://doi.org/10.1186/s40795-025-01007-3}.

\bibitem{who_sodium}
World Health Organization, \emph{Sodium Intake for Adults and Children}. Geneva, Switzerland: World Health Organization, 2012. Available: \url{https://www.who.int/publications/i/item/9789241504836}

\bibitem{r214}
Department of Health, South Africa, ``Regulations relating to the reduction of sodium in certain foodstuffs and related matters,'' Government Notice R.214, Government Gazette No. 36274, 2013. Available: \url{https://www.gov.za/sites/default/files/gcis_document/201409/36274rg9934gon214.pdf}

\bibitem{charlton_compliance}
K. E. Charlton, B. Pretorius, R. Shakhane, P. Naidoo, H. Cimring, K. Hussain, B. Nojilana, and J. Webster, ``Compliance of the food industry with mandated salt target levels in South Africa: towards development of a monitoring and surveillance framework,'' \emph{Journal of Food Composition and Analysis}, vol. 126, Art. no. 105908, pp. 1--7, 2024, doi: \url{https://doi.org/10.1016/j.jfca.2023.105908}.

\bibitem{vdw_compliance}
B. van der Westhuizen, T. Frank, S. Abdool Karim, and E. C. Swart, ``Determining food industry compliance to mandatory sodium limits: successes and challenges from the South African experience,'' \emph{Public Health Nutrition}, vol. 26, no. 11, pp. 2551--2558, 2023, doi: \url{https://doi.org/10.1017/S1368980023000757}.

\bibitem{peters_sodium}
S. A. E. Peters, E. Dunford, L. J. Ware, T. Harris, A. Walker, M. Wicks, T. van Zyl, B. Swanepoel, K. E. Charlton, M. Woodward, J. Webster, and B. Neal, ``The sodium content of processed foods in South Africa during the introduction of mandatory sodium limits,'' \emph{Nutrients}, vol. 9, no. 4, Art. no. 404, 2017, doi: \url{https://doi.org/10.3390/nu9040404}.

\bibitem{swanepoel_sodium}
B. Swanepoel, L. Malan, P. H. Myburgh, R. Schutte, K. Steyn, and E. Wentzel-Viljoen, ``Sodium content of foodstuffs included in the sodium reduction regulation of South Africa,'' \emph{Journal of Food Composition and Analysis}, vol. 63, pp. 73--78, 2017, doi: \url{https://doi.org/10.1016/j.jfca.2017.07.040}.

\bibitem{assiri_llm}
F. Y. Assiri, M. D. Alahmadi, M. A. Almuashi, and A. M. Almansour, ``Extract nutritional information from bilingual food labels using large language models,'' \emph{Journal of Imaging}, vol. 11, no. 8, Art. no. 271, 2025, doi: \url{https://doi.org/10.3390/jimaging11080271}.

\bibitem{pettersson_grocery}
T. Pettersson, M. Riveiro, and T. L\"{o}fstr\"{o}m, ``Multimodal fine-grained grocery product recognition using image and OCR text,'' \emph{Machine Vision and Applications}, vol. 35, Art. no. 79, 2024, doi: \url{https://doi.org/10.1007/s00138-024-01549-9}.

\bibitem{ma_vlm}
P. Ma, Y. Wu, N. Yu, Y. Zhang, M. Backes, Q. Wang, and C. Wei, ``Integrating Vision-Language Models for Accelerated High-Throughput Nutrition Screening,'' \emph{Advanced Science}, vol. 11, no. 34, Art. no. e2403578, 2024, doi: \url{https://doi.org/10.1002/advs.202403578}.

\bibitem{qwen25vl}
S. Bai et al., ``Qwen2.5-VL Technical Report,'' \emph{arXiv preprint arXiv:2502.13923}, 2025. Available: https://arxiv.org/abs/2502.13923

\bibitem{ultralytics_yolo}
G. Jocher, J. Qiu, and A. Chaurasia, ``Ultralytics YOLO,'' Zenodo, version 8.4.47, May 2026, doi: \url{https://doi.org/10.5281/zenodo.20052427}.

\bibitem{microsoft_vott}
Microsoft, ``Visual Object Tagging Tool (VoTT),'' GitHub repository. Available: \url{https://github.com/microsoft/VoTT}

\bibitem{who_sodium_reduction}
World Health Organization, ``Sodium reduction,'' Fact sheet. Available: \url{https://www.who.int/news-room/fact-sheets/detail/sodium-reduction}

\bibitem{groq_api}
Groq, ``Groq API Reference,'' Groq Docs. Available: https://console.groq.com/docs/api-reference

\end{thebibliography}
\end{document}